\documentclass[11pt,a4paper]{article}
\usepackage[hyphens]{url}
\usepackage[hyperref]{acl2017}
\usepackage{times}
\usepackage{latexsym}
\usepackage{amsmath,amssymb}
\usepackage{multirow}
\usepackage{color}
\usepackage{graphicx}
\usepackage{bbm}
\usepackage{xspace}
\usepackage{wasysym}
\usepackage{algorithmic}
\usepackage{float}
\usepackage{mathtools}
\usepackage{pgfplots}
\usepackage{placeins}
\usepackage{tabularx}
\usepgfplotslibrary{groupplots}
\pgfplotsset{compat=1.14}
\aclfinalcopy 

\newcolumntype{L}[1]{>{\raggedright\arraybackslash}p{#1}}
\newcolumntype{C}[1]{>{\centering\arraybackslash}p{#1}}
\newcolumntype{R}[1]{>{\raggedleft\arraybackslash}p{#1}}\mathtoolsset{showonlyrefs}

\newcommand{\jwcomment}[1]{}

\newcommand{\sent}{s}
\newcommand{\sentt}{s_t}

\newcommand{\paragramsl}{\textsc{paragram-sl999}\xspace}

\newcommand{\ganlong}{\textsc{gated recurrent averaging network}\xspace}
\newcommand{\gan}{\textsc{GRAN}\xspace}
\newcommand{\lstmavg}{LSTM\textsc{avg}\xspace}
\newcommand{\wordavg}{\textsc{avg}\xspace}
\newcommand{\simpwiki}{SimpWiki\xspace}

\newcommand\norm[1]{\left\lVert#1\right\rVert}
\DeclareMathOperator*{\argmax}{argmax}

\title{Revisiting Recurrent Networks for Paraphrastic Sentence Embeddings}

\author{John Wieting \ \ \ \ \ \  Kevin Gimpel\\
Toyota Technological Institute at Chicago, Chicago, IL, 60637, USA\\
\tt{\{jwieting,kgimpel\}@ttic.edu}
}

\date{}

\begin{document}
\maketitle
\begin{abstract}
We consider the problem of learning general-purpose, paraphrastic sentence embeddings, revisiting the setting of \newcite{wieting-16-full}. While they found LSTM recurrent networks to underperform word averaging, 
we present several developments that together produce the opposite conclusion. 
These include 
training on sentence pairs rather than phrase pairs, 
averaging states to represent sequences, and regularizing aggressively. 
These improve LSTMs in both transfer learning and supervised settings. We also introduce a new recurrent architecture, the \ganlong, that is inspired by averaging and LSTMs while outperforming them both. 
We analyze our learned models, finding evidence of preferences for particular parts of speech and dependency relations. 
\footnote{Trained models and code are available at \url{http://ttic.uchicago.edu/~wieting}.}
\end{abstract}

\section{Introduction}
Modeling sentential compositionality is a fundamental aspect of natural language semantics. Researchers have proposed a broad range of compositional functional architectures~\cite{mitchell2008vector,SocherEtAl2011:PoolRAE,kalchbrenner-grefenstette-blunsom:2014:P14-1} and evaluated them on a large variety of applications. Our goal is to learn a general-purpose sentence embedding function that can be used unmodified for measuring semantic textual similarity (STS) \cite{agirre2012semeval} and can also serve as a useful initialization for downstream tasks. We wish to learn this embedding function such that sentences with high semantic similarity have high cosine similarity in the embedding space. In particular, we focus on the setting of \newcite{wieting-16-full}, in which models are trained on noisy paraphrase pairs and evaluated on both STS and supervised semantic tasks. 

Surprisingly, Wieting et al.~found that simple embedding functions\----those based on averaging word vectors\----outperform more powerful architectures based on long short-term memory (LSTM)~\cite{hochreiter1997long}. In this paper, we revisit their experimental setting and present several techniques that together improve the performance of the LSTM to be superior to word averaging. 

We first change data sources: rather than train on noisy phrase pairs from the Paraphrase Database (PPDB; Ganitkevitch et al., 2013),\nocite{GanitkevitchDC13} we use noisy \emph{sentence} pairs obtained automatically by aligning Simple English to standard English Wikipedia \cite{coster2011simple}. Even though this data was intended for use by text simplification systems, we find it to be efficient and effective for learning sentence embeddings, outperforming much larger sets of examples from PPDB. 

We then show how we can modify and regularize the LSTM to further improve its performance. The main modification is to simply average the hidden states instead of using the final one. For regularization, we experiment with two kinds of dropout and also with randomly 
scrambling
the words in each input sequence. We find that these techniques help in the transfer learning setting and on two supervised semantic similarity datasets as well. Further gains are obtained on the supervised tasks by initializing with our models from the transfer setting.  

Inspired by the strong performance of both averaging and LSTMs, we introduce a novel recurrent neural network architecture which we call the \ganlong (\gan). 
The \gan outperforms averaging and the LSTM in both the transfer and supervised learning settings, forming a promising new recurrent architecture for semantic modeling.

\section{Related Work}
\label{sec:relwork}
Modeling sentential compositionality has received a great deal of attention in recent years. A comprehensive survey is beyond the scope of this paper, but we mention popular functional families: 
neural bag-of-words models~\cite{kalchbrenner-grefenstette-blunsom:2014:P14-1}, deep averaging networks (DANs)~\cite{iyyer-EtAl:2015:ACL-IJCNLP}, 
recursive neural networks using syntactic parses~\cite{SocherEtAl2011:PoolRAE,socher-EtAl:2012:EMNLP-CoNLL,socher-13,irsoy-drsv}, 
convolutional neural networks~\cite{kalchbrenner-grefenstette-blunsom:2014:P14-1,kim-14,NIPS20145550}, 
and recurrent neural networks using long short-term memory~\cite{tai2015improved,ling-EtAl:2015:EMNLP2,liu-EtAl:2015:EMNLP2}. 
Simple operations based on vector addition and multiplication typically serve as strong baselines~\cite{mitchell2008vector,Mitchell:Lapata:2010,Blacoe2012}. 

Most work cited above uses a supervised learning framework, so the composition function is learned  discriminatively for a particular task. 
In this paper, we are primarily interested in creating general purpose, domain independent embeddings for word
sequences. Several others have pursued this goal~\citep{SocherEtAl2011:PoolRAE,le2014distributed,pham-EtAl:2015:ACL-IJCNLP,kiros2015skip,hill2016learning,arora2017simple,pgj2017unsup}, though usually with the intent to extract useful features for supervised sentence tasks rather than to capture semantic similarity. 

An exception is the work of \newcite{wieting-16-full}. We closely follow their experimental setup and directly address some outstanding questions in their experimental results. 
Here we briefly summarize their main findings and their attempts at explaining them. They made the surprising discovery that word averaging outperforms LSTMs by a wide margin in the transfer learning setting. They proposed several hypotheses for why this occurs. They first considered that the LSTM was unable to adapt to the differences in sequence length between phrases in training and sentences in test.
This was ruled out by showing that neither model showed any strong correlation between sequence length and performance on the test data. 

They next examined whether the LSTM was overfitting on the training data, but then showed that both models achieve similar values of the training objective and similar performance on \emph{in-domain} held-out test sets. 
Lastly, they considered whether their hyperparameters were inadequately tuned, but extensive hyperparameter tuning did not change the story. 
Therefore, the reason for the performance gap, and how to correct it, was left as an open problem. This paper takes steps toward addressing that problem.

\section{Models and Training} 
\subsection{Models}

Our goal is to embed a word sequence $\sent$ into a fixed-length vector. We focus on three compositional models in this paper, all of which use words as the smallest unit of compositionality. We denote the $t$th word in $\sent$ as $\sentt$, and we denote its word embedding by $x_t$. 

Our first two models have been well-studied in prior work, so we  describe them briefly. The first, which we call \wordavg, simply averages the embeddings $x_t$ of all words in $\sent$. The only parameters learned in this model are those in the word embeddings themselves, which are stored in the word embedding matrix $W_w$.  
This model was found by \newcite{wieting-16-full} to perform very strongly for semantic similarity tasks. 

Our second model uses a long short-term memory (LSTM) recurrent neural network~\citep{hochreiter1997long} to embed $\sent$. 
We use the LSTM variant from \newcite{gers2003learning} including its ``peephole'' connections. We consider two ways to obtain a sentence embedding from the LSTM. The first uses the final hidden vector, which we denote $h_{-1}$. The second, denoted \lstmavg, averages all hidden vectors of the LSTM. In both variants, the learnable parameters include both the LSTM parameters $W_c$ and the word embeddings $W_w$. 

Inspired by the success of the two models above, we propose a third model, which we call the \ganlong (\gan). The \ganlong combines the benefits of \wordavg and LSTMs. In fact it reduces to \wordavg if the output of the gate is all ones. We first use an LSTM to generate a hidden vector, $h_t$, for each word $\sentt$ in $\sent$. Then we use $h_t$ to compute a gate that will be elementwise-multiplied with $x_t$, resulting in a new, gated hidden vector $a_t$ for each step $t$:
\begin{equation}
a_t = x_t\odot\sigma(W_xx_t + W_hh_t + b) \label{eq:gan}
\end{equation}
\noindent 
where $W_x$ and $W_h$ are parameter matrices, $b$ is a parameter vector, and $\sigma$ is the elementwise logistic sigmoid function. After all $a_t$ have been generated for a sentence, they are averaged to produce the embedding for that sentence. This model includes as learnable parameters those of the LSTM, the word embeddings, and the additional parameters in Eq.~\eqref{eq:gan}. For both the LSTM and \gan models, we use $W_c$ to denote the ``compositional'' parameters, i.e., all parameters other than the word embeddings. 

The motivation for the \gan is that we are contextualizing the word embeddings prior to averaging. The gate can be seen as an attention, attending to the prior context of the sentence.\footnote{We tried a variant of this model without the gate. We obtain $a_t$ from $f(W_xx_t + W_hh_t + b)$, where $f$ is a nonlinearity, tuned over tanh and ReLU. The performance of the model is significantly worse than the \gan in all experiments.}

We also experiment with four other variations of this model, though they generally were more complex and showed inferior performance. 
In the first, \gan-2, the gate is applied to $h_t$ (rather than $x_t$) to produce $a_t$, and then these $a_t$ are averaged as before. 

\gan-3 and \gan-4 use two gates: one applied to $x_t$ and one applied to $a_{t-1}$. We tried two different ways of computing these gates: for each gate $i$, $\sigma(W_{x_i}x_t + W_{h_i}h_t + b_i)$ (\gan-3) or $\sigma(W_{x_i}x_t + W_{h_i}h_t + W_{a_i}a_{t-1} + b_i)$ (\gan-4). The sum of these two terms comprised $a_t$. In this model, the last average hidden state, $a_{-1}$, was used as the sentence embedding after dividing it by the length of the sequence. In these models, we are additionally keeping a running average of the embeddings that is being modified by the context at every time step. In \gan-4, this running average is also considered when producing the contextualized word embedding.

Lastly, we experimented with a fifth \gan, \gan-5, in which we use two gates, calculated by $\sigma(W_{x_i}x_t + W_{h_i}h_t + b_i)$ for each gate $i$. The first is applied to $x_t$ and the second is applied to $h_t$. The output of these gates is then summed. Therefore \gan-5 can be reduced to either word-averaging or averaging LSTM states, depending on the behavior of the gates. If the first gate is all ones and the second all zeros throughout the sequence, the model is equivalent to word-averaging. Conversely, if the first gate is all zeros and the second is all ones throughout the sequence, the model is equivalent to averaging the LSTM states. Further analysis of these models is included in Section~\ref{sec:exp}.

\subsection{Training}

We follow the training procedure of \newcite{wieting2015ppdb-short} and \newcite{wieting-16-full}, described below. The training data consists of a set $S$ of phrase or sentence pairs $\langle s_1, s_2\rangle$ from either the Paraphrase Database (PPDB; Ganitkevitch et al., 2013)\nocite{GanitkevitchDC13} or the aligned Wikipedia sentences~\cite{coster2011simple} where $s_1$ and $s_2$ are assumed to be paraphrases. We optimize a margin-based loss: 

\begin{small}
\begin{align}
&\underset{W_c,W_w}{\text{min}} \frac{1}{|S|}\Bigg(\sum_{\langle s_1,s_2\rangle \in S} 
\max(0,\delta - \cos(g(s_1), g(s_2))\\
&+ \cos(g(s_1), g(t_1))) + \max(0,\delta - \cos(g(s_1),g(s_2))\\
&+ \cos(g(s_2), g(t_2)))\bigg) + \lambda_c\norm{W_c}^2  + \lambda_{w}\norm{W_{w_{\mathit{initial}}} - W_w}^2
\label{eq:obj}
\end{align} 
\end{small}

\noindent where $g$ is the model in use (e.g., \wordavg or LSTM), $\delta$ is the margin, $\lambda_{c}$ and $\lambda_{w}$ are regularization parameters, $W_{w_{\mathit{initial}}}$ is the initial word embedding matrix, and $t_1$ and $t_2$ are carefully-selected negative examples taken from a mini-batch during optimization.  
The intuition is that we want the two phrases to be more similar to each other ($\cos(g(s_1), g(s_2))$) than either is to their respective negative examples $t_1$ and $t_2$, by a margin of at least $\delta$. 

\subsubsection{Selecting Negative Examples}
\label{sec:sample}
To select $t_1$ and $t_2$ in Eq.~\eqref{eq:obj}, we simply choose the most similar phrase in some set of phrases (other than those in the given phrase pair). 
For simplicity 
we use the mini-batch for this set, but it could be a different set. That is, we choose $t_1$ for a given $\langle s_1, s_2\rangle$ as follows:
\begin{equation}
t_1 = \argmax_{t : \langle t, \cdot\rangle \in S_b \setminus \{\langle s_1, s_2\rangle\}} \cos(g(s_1), g(t))
\end{equation}
\noindent where $S_b\subseteq S$ is the current mini-batch. 
That is, we want to choose a negative example $t_i$ that is similar to $s_i$ according to the current model. 
The downside is that we may occasionally choose a phrase $t_i$ that is actually a true paraphrase of $s_i$.

\section{Experiments}  \label{sec:exp}
Our experiments are designed to address the empirical question posed by \newcite{wieting-16-full}: why do LSTMs underperform \wordavg for transfer learning? 
In Sections~\ref{sec:exp:data}-\ref{sec:exp:reg}, we make progress on this question by presenting methods that bridge the gap between the two models in the transfer setting. 
We then apply these same techniques to improve performance in the supervised setting, described in Section~\ref{sec:exp:sup}. In both settings we also evaluate our novel \gan architecture, finding it to consistently outperform both \wordavg and the LSTM. 

\subsection{Transfer Learning}

\subsubsection{Datasets and Tasks}
We train on large sets of noisy paraphrase pairs and evaluate on a diverse set of 22 textual similarity datasets, including all datasets from every SemEval semantic textual similarity (STS) task from 2012 to 2015. We also evaluate on the SemEval 2015 Twitter task \cite{xu2015semeval} and the SemEval 2014 SICK Semantic Relatedness task \cite{marelli2014semeval}. 
Given two sentences, the aim of the STS tasks is to predict their similarity on a 0-5 scale, where 0 indicates the sentences are on different topics and 5 indicates that they are completely equivalent. 
We report the average Pearson's $r$ over these 22 sentence similarity tasks. 

Each STS task consists of 4-6 datasets covering a wide variety of domains, including newswire, tweets, glosses, machine translation outputs, web forums, news headlines, image and video captions, among others. 
Further details are provided in the official task descriptions~\cite{agirre2012semeval,diab2013eneko,agirre2014semeval,agirre2015semeval}.

\subsubsection{Experiments with Data Sources}
\label{sec:exp:data}
We first investigate how different sources of training data affect the results. We try two data sources. The first is phrase pairs from the Paraphrase Database (PPDB). PPDB comes in different sizes (S, M, L, XL, XXL, and XXXL), where each larger size subsumes all smaller ones. The pairs in PPDB are sorted by a confidence measure and so the smaller sets contain higher precision paraphrases. PPDB is derived automatically from naturally-occurring bilingual text, and versions of PPDB have been released for many languages without the need for any manual annotation \cite{ganitkevitch2014multilingual}. 

The second source of data is a set of sentence pairs automatically extracted from Simple English Wikipedia and English Wikipedia articles by \newcite{coster2011simple}. This data was extracted for developing text simplification systems, where each instance pairs a simple and complex sentence representing approximately the same information. Though the data was obtained for simplification, we use it as a source of training data for learning paraphrastic sentence embeddings. The dataset, which we call \simpwiki, consists of 167,689 sentence pairs. 

To ensure a fair comparison, we select a sample of pairs from PPDB XL such that the number of tokens is approximately the same as the number of tokens in the \simpwiki sentences.\footnote{The PPDB data consists of 1,341,188 phrase pairs and contains 3 more tokens than the \simpwiki data.} 

We use \paragramsl embeddings~\cite{wieting2015ppdb-short} to initialize the word embedding matrix ($W_w$) for all models. For all experiments, we fix the mini-batch size to 100, and $\lambda_{c}$ to 0. We tune the margin $\delta$ over $\{0.4,0.6,0.8\}$ and $\lambda_{w}$ over $\{10^{-4},10^{-5},10^{-6},10^{-7},10^{-8},0\}$. We train \wordavg for 7 epochs, and the LSTM for 3, since it converges much faster and does not benefit from 7 epochs. For optimization we use Adam~\cite{kingma2014adam} with a learning rate of 0.001. 
We use the 2016 STS tasks~\cite{agirre2016semeval} for model selection, where we average the Pearson's $r$ over its 5 datasets. We refer to this type of model selection as {\it test}. For evaluation, we report the average Pearson's $r$ over the 22 other sentence similarity tasks.

\begin{table}[t]
\setlength{\tabcolsep}{4pt}
\centering
\begin{tabular} {| l | c | c |c|}
\cline{2-4}
\multicolumn{1}{c|}{} & \wordavg & LSTM & \lstmavg \\
\hline
PPDB & 67.7 & 54.2 & 64.2\\
\hline
\simpwiki & 68.4 & 59.3 & 67.5 \\
\hline
\end{tabular}
\caption{\label{table:data} 
Test results on SemEval semantic textual similarity datasets (Pearson's $r \times 100$) when training on different sources of data: phrase pairs from PPDB or simple-to-standard English Wikipedia sentence pairs from \newcite{coster2011simple}. 
}
\end{table}

The results are shown in Table~\ref{table:data}. 
We first note that, when training on PPDB, we find the same result as \newcite{wieting-16-full}: \wordavg outperforms the LSTM by more than 13 points. However, when training both on sentence pairs, the gap shrinks to about 9 points. It appears that part of the inferior performance for the LSTM in prior work was due to training on phrase pairs rather than on sentence pairs. The \wordavg model also benefits from training on sentences, but not nearly as much as the LSTM.\footnote{We experimented with adding EOS tags at the end of training and test sentences, SOS tags at the start of training and test sentences, adding both, and adding neither. We treated adding these tags as hyperparameters and tuned over these four settings along with the other hyperparameters in the original experiment. Interestingly, we found that adding these tags, especially EOS, had a large effect on the LSTM when training on \simpwiki, improving performance by 6 points. When training on PPDB, adding EOS tags only improved performance by 1.6 points.

The addition of the tags had a smaller effect on \lstmavg. Adding EOS tags improved performance by 0.3 points on \simpwiki and adding SOS tags on PPDB improved performance by 0.9 points. 
}

Our hypothesis explaining this result is that in PPDB, the “phrase pairs” are short fragments of text which are not necessarily constituents or phrases in any syntactic sense. Therefore, the sentences in the STS test sets are quite different from the fragments seen during training. We hypothesize that while word-averaging is relatively unaffected by this difference, the recurrent models are much more sensitive to overall characteristics of the word sequences, and the difference between train and test matters much more. 

These results also suggest that the \simpwiki data, even though it was developed for text simplification, may be useful for other researchers working on semantic textual similarity tasks. 

\subsubsection{Experiments with LSTM Variations}
\label{sec:exp:vari}
We next compare LSTM and \lstmavg. The latter consists of averaging the hidden vectors of the LSTM rather than using the final hidden vector as in prior work~\cite{wieting-16-full}. 
We hypothesize that the LSTM may put more emphasis on the words at the end of the sentence than those at the beginning. By averaging the hidden states, the impact of all words in the sequence is better taken into account. Averaging also makes the LSTM more like \wordavg, which we know to perform strongly in this setting. 

The results on \wordavg and the LSTM models are shown in Table~\ref{table:data}. 
When training on PPDB, moving from LSTM to \lstmavg improves performance by 10 points, closing most of the gap with \wordavg. 
We also find that \lstmavg improves by moving from PPDB to \simpwiki, though in both cases it still lags behind \wordavg.  
\subsection{Experiments with Regularization}
\label{sec:exp:reg}
We next experiment with various forms of regularization.   
Previous work \cite{wieting-16-full,wieting2016charagram} only used L$_2$ regularization. \newcite{wieting-16-full} also regularized the word embeddings back to their initial values. Here we use L$_2$ regularization as well as several additional regularization methods we describe below. 

We try two forms of dropout. The first is just standard dropout~\cite{srivastava2014dropout} on the word embeddings. The second is ``word dropout'', which drops out entire word embeddings with some probability~\cite{iyyer-EtAl:2015:ACL-IJCNLP}. 

We also experiment with scrambling 
the inputs. For a given mini-batch, we 
go through each sentence pair and, with some probability, we shuffle the words in each sentence in the pair. When scrambling a sentence pair, we always shuffle both sentences in the pair. We do this before selecting negative examples for the mini-batch. The motivation for scrambling is to make it more difficult for the LSTM to memorize the sequences in the training data, forcing it to focus more on the identities of the words and less on word order. Hence it will be expected to behave more like the word averaging model.\footnote{We also tried some variations on scrambling that did not yield significant improvements: scrambling after obtaining the negative examples, partially scrambling by performing $n$ swaps where $n$ comes from a Poisson distribution with a tunable $\lambda$, and scrambling individual sentences with some probability instead of always scrambling both in the pair.} 

We also experiment with combining scrambling and dropout. In this setting, we tune over scrambling with either word dropout or dropout.

The settings for these experiments are largely the same as those of the previous section with the exception that we tune $\lambda_w$ over a smaller set of values: $\{10^{-5},0\}$. When using L$_2$ regularization, we tune $\lambda_c$ over $\{10^{-3},10^{-4},10^{-5},10^{-6}\}$. When using dropout, we tune the dropout rate over $\{0.2,0.4,0.6\}$. When using scrambling, we tune the scrambling rate over $\{0.25,0.5,0.75\}$. We also include a bidirectional model (``Bi'') for both \lstmavg and the \ganlong. We tune over two ways to combine the forward and backward hidden states; the first simply adds them together and the second uses a single feedforward layer with a $\tanh$ activation. 

We try two approaches for model selection. The first, {\it test}, is the same as was done in Section~\ref{sec:exp:data}, where we use the average Pearson's $r$ on the 5 2016 STS datasets. The second tunes based on the average Pearson's $r$ of all 22 datasets in our evaluation. We refer to this as {\it oracle}.

\begin{table}[t]
\setlength{\tabcolsep}{4pt}
\small
\centering
\begin{tabular} {| l |l| c | c |} \hline
Model & Regularization & Oracle & 2016 STS \\
\hline
\multirow{3}{*}{\wordavg} & none & 68.5 & 68.4 \\
 & dropout & 68.4 & 68.3 \\
 & word dropout & 68.3 & 68.3 \\
\hline
 \multirow{5}{*}{LSTM} & none & 60.6 & 59.3\\
 & L$_2$ & 60.3 & 56.5\\
 & dropout & 58.1 & 55.3\\
 & word dropout & 66.2 & 65.3\\
 & scrambling & 66.3 & 65.1\\
 & dropout, scrambling & 68.4 & 68.4\\
\hline
\multirow{2}{*}{\lstmavg} & none & 67.7 & 67.5\\
& dropout, scrambling & 69.2 & 68.6\\
\hline
Bi\lstmavg & dropout, scrambling & \bf 69.4 & \bf 68.7\\
\hline
\end{tabular}
\caption{\label{table:transfer} 
Results on SemEval textual similarity datasets (Pearson's $r \times 100$) when experimenting with different regularization techniques. 
}
\end{table}

\begin{table}[t]
\setlength{\tabcolsep}{4pt}
\small
\centering
\begin{tabular} {|l|c|c|}
\hline
Model & Oracle & STS 2016 \\
\hline
\gan (no reg.) & 68.0 & 68.0 \\
\gan & 69.5 & \bf 68.9 \\
\gan-2 & 68.8 & 68.1 \\
\gan-3 & 69.0 & 67.2 \\
\gan-4 & 68.6 & 68.1 \\
\gan-5 & 66.1 & 64.8 \\
Bi\gan & \bf 69.7 & 68.4\\
\hline
\end{tabular}
\caption{\label{table:gan} 
Results on SemEval textual similarity datasets (Pearson's $r \times 100$) for the \gan architectures. The first row, marked as (no reg.) is the \gan without any regularization. The other rows show the result of the various \gan models using dropout and scrambling.
}
\end{table}

\begin{table}
\setlength{\tabcolsep}{4pt}
\small
\centering
\begin{tabular} { | l || C{1.3cm} | C{1.3cm} | C{1.3cm} |} 
\hline
Dataset & \lstmavg & \wordavg & \gan \\
\hline
MSRpar & \bf 49.0 & 45.9 & 47.7\\
MSRvid &  84.3 & 85.1 & \bf 85.2\\
SMT-eur & \bf 51.2 & 47.5 & 49.3\\
OnWN & \bf 71.5 & 71.2 & 71.5\\
SMT-news & \bf 68.0 & 58.2 & 58.7\\
\hline
STS 2012 Average & \bf 64.8 & 61.6 & 62.5\\
\hline
headline & \bf 77.3 & 76.9 & 76.1\\
OnWN & 81.2 & 72.8 & \bf 81.4\\
FNWN & 53.2 & 50.2 & \bf 55.6\\
SMT & \bf 40.7 & 38.0 & 40.3\\
\hline
STS 2013 Average & 63.1 & 59.4 & \bf 63.4\\
\hline
deft forum & \bf 56.6 & 55.6 & 55.7\\
deft news & 78.0 & \bf 78.5 & 77.1\\
headline & 74.5 & \bf 75.1 & 72.8\\
images & 84.7 & 85.6 & \bf 85.8\\
OnWN & 84.9 & 81.4 & \bf 85.1\\
tweet news & 76.3 & \bf 78.7 & 78.7\\
\hline
STS 2014 Average & 75.8 & 75.8 & \bf 75.9\\
\hline
answers-forums & 71.8 & 70.6 & \bf 73.1\\
answers-students & 71.1 & \bf 75.8 & 72.9\\
belief & 75.3 & 76.8 & \bf 78.0\\
headline & 79.5 & \bf 80.3 & 78.6\\
images & 85.8 & \bf 86.0 & 85.8\\
\hline
STS 2015 Average & 76.7 & \bf 77.9 & 77.7\\
\hline
2014 SICK & 71.3 & 72.4 & \bf 72.9\\
\hline
2015 Twitter & \bf 52.1 & \bf 52.1 & 50.2\\
\hline
\end{tabular}
\caption{\label{table:fullresults}
Results on SemEval textual similarity datasets (Pearson's $r \times 100$). The highest score in each row is in boldface. 
}
\vspace{-0.2cm}
\end{table}

The results are shown in Table~\ref{table:transfer}. They show that dropping entire word embeddings and scrambling input sequences is very effective in improving the result of the LSTM, while neither type of dropout improves \wordavg. Moreover, averaging the hidden states of the LSTM is the most effective modification to the LSTM in improving performance. All of these modifications can be combined to significantly improve the LSTM, finally allowing it to overtake \wordavg. 

In Table~\ref{table:gan}, we compare the various \gan architectures. We find that the \gan provides a small improvement over the best LSTM configuration, possibly because of its similarity to \wordavg. It also outperforms the other \gan models, despite being the simplest.  

In Table~\ref{table:fullresults}, we show results on all individual STS evaluation datasets after using STS 2016 for model selection (unidirectional models only). The \lstmavg and \ganlong are more closely correlated in performance, in terms of Spearman's $\rho$ and Pearson'r $r$, than either is to \wordavg. But they do differ significantly in some datasets, most notably in those comparing machine translation output with its reference. Interestingly, both the \lstmavg and \ganlong significantly outperform \wordavg in the datasets focused on comparing glosses like {\it OnWN} and {\it FNWN}. Upon examination, we found that these datasets, especially 2013 {\it OnWN}, contain examples of low similarity with high word overlap. For example, the pair $\langle${\it the act of preserving or protecting something.}, {\it the act of decreasing or reducing something.}$\rangle$ from 2013 {\it OnWN} has a gold similarity score of 0.4. It appears that \wordavg was fooled by the high amount of word overlap in such pairs, while the other two models were better able to recognize the semantic differences. 

\subsection{Supervised Text Similarity}
\label{sec:exp:sup}

We also investigate if these techniques can improve LSTM performance on supervised semantic textual similarity tasks. We evaluate on two supervised datasets. For the first, we start with the 20 SemEval STS datasets from 2012-2015 and then use 40\% of each dataset for training, 10\% for validation, and the remaining 50\% for testing. There are 4,481 examples in training, 1,207 in validation, and 6,060 in the test set. 
The second is the SICK 2014 dataset, using its standard training, validation, and test sets. There are 4,500 sentence pairs in the training set, 500 in the development set, and 4,927 in the test set. 
The SICK task is an easier learning problem since the training examples are all drawn from the same distribution, and they are mostly shorter and use simpler language. As these are supervised tasks, the sentence pairs in the training set contain manually-annotated semantic similarity scores. 

We minimize the loss function\footnote{This objective function has been shown to perform very strongly on text similarity tasks, significantly better than squared or absolute error.} from \newcite{tai2015improved}. Given a score for a sentence pair in the range $[1, K]$, where $K$ is an integer, with sentence representations $h_L$ and $h_R$, and model parameters $\theta$, they first compute:
\begin{align}
h_\times &= h_L \odot h_R, \label{eq:objsup1} \:\:h_+ = |h_L - h_R|, \nonumber \\
h_s &= \sigma\left(W^{(\times)} h_\times  + W^{(+)} h_+ + b^{(h)} \right), \nonumber \\
\hat{p}_\theta     &= \mathrm{softmax}\left(W^{(p)} h_s + b^{(p)} \right), \nonumber \\
\hat{y}     &= r^T \hat{p}_\theta, \nonumber
\end{align}
where $r^T = [1~2~\dots~K]$. They then define a sparse target distribution $p$ that satisfies $y = r^T p$:
\begin{equation}
p_i = \begin{cases} y - \lfloor y \rfloor, & i = \lfloor y \rfloor + 1 \\ \lfloor y \rfloor - y + 1, & i = \lfloor y \rfloor  \\ 0 & \text{otherwise}\end{cases}
\end{equation}
for $1 \leq i \leq K$. Then they use the following loss, the regularized KL-divergence between $p$ and $\hat{p}_\theta$:
\begin{equation}
 J(\theta) = \frac{1}{m} \sum_{k=1}^m \mathrm{KL}\Big(p^{(k)}~\Big\|~\hat{p}^{(k)}_\theta\Big),\label{eq:objsup}
\end{equation}
where $m$ is the number of training pairs. 

We experiment with the LSTM, \lstmavg, and \wordavg models with dropout, word dropout, and scrambling tuning over the same hyperparameter as in Section~\ref{sec:exp:reg}. We again regularize the word embeddings back to their initial state, tuning $\lambda_w$ over $\{10^{-5},0\}$. We used the validation set for each respective dataset for model selection. 

\begin{table}[t]
\setlength{\tabcolsep}{4pt}
\small
\centering
\begin{tabular} {| l | l | c | c | c |} \hline
Model & Regularization & STS & SICK & Avg. \\
\hline
\multirow{4}{*}{\wordavg}
 & none & 79.2 & 85.2  & 82.2  \\
 & dropout & 80.7 & 84.5  & 82.6  \\
 & word dropout & 79.3 & 81.8  & 80.6  \\
\hline
 & none & 68.4 & 80.9 & 74.7 \\
 & dropout & 69.6 & 81.3 & 75.5 \\
LSTM & word dropout & 68.0 & 76.4 & 72.2 \\
 & scrambling & 74.2 & 84.4 & 79.3 \\
 & dropout, scrambling & 75.0 & 84.2 & 79.6 \\
\hline
\multirow{5}{*}{\lstmavg} &  none & 69.0 & 79.5 & 74.3 \\
 & dropout & 69.2 & 79.4 & 74.3 \\
 & word dropout & 65.6 & 76.1 & 70.9 \\
 & scrambling & 76.5 & 83.2 & 79.9 \\
 & dropout, scrambling & 76.5 & 84.0 & 80.3 \\
\hline
\multirow{5}{*}{\gan} & none & 79.7 & 85.2 & 82.5 \\
 & dropout & 79.7 & 84.6 & 82.2 \\
 & word dropout & 77.3 & 83.0 & 80.2 \\
 & scrambling & 81.4 & \bf 85.3 & \bf 83.4 \\
 & dropout, scrambling & \bf 81.6 & 85.1 & \bf 83.4 \\
\hline
\end{tabular}
\caption{\label{table:supervised} 
Results from supervised training on the STS and SICK datasets (Pearson's $r \times 100$). The last column is the average result on the two datasets.
}
\end{table}

\begin{table}[t]
\setlength{\tabcolsep}{4pt}
\small
\centering
\begin{tabular} {|l|c|c|c|}
\hline
Model & STS & SICK & Avg. \\
\hline
\gan & \bf 81.6 & 85.3 & \bf 83.5 \\
\gan-2 & 77.4 & 85.1 & 81.3 \\
\gan-3 & 81.3 & 85.4 & 83.4\\
\gan-4 & 80.1 & \bf 85.5 & 82.8\\
\gan-5 & 70.9 & 83.0 & 77.0\\
\hline
\end{tabular}
\caption{\label{table:gansupervised} 
Results from supervised training on the STS and SICK datasets (Pearson's $r \times 100$) for the \gan architectures. The last column is the average result on the two datasets. 
}
\end{table}

The results are shown in Table~\ref{table:supervised}. 
The \ganlong has the best performance on both datasets. Dropout helps the word-averaging model in the STS task, unlike in the transfer learning setting. The LSTM benefits slightly from dropout, scrambling, and averaging on their own individually with the exception of word dropout on both datasets and averaging on the SICK dataset. However, when combined, these modifications are able to significantly improve the performance of the LSTM, bringing it much closer in performance to \wordavg. This experiment indicates that these modifications when training LSTMs are beneficial outside the transfer learning setting, and can potentially be used to improve performance for the broad range of problems that use LSTMs to model sentences. 

In Table~\ref{table:gansupervised} we compare the various \gan architectures under the same settings as the previous experiment. We find that the \gan still has the best overall performance.

\begin{table*}[th!]
\setlength{\tabcolsep}{4pt} 
\small
\centering
\begin{tabular} { | c | p{5.4cm}  p{6.8cm} | c | c | c |} \hline
\# & Sentence 1 & Sentence 2 &  \textsc{Lavg} & \wordavg & Gold \\\hline
1 & the lamb is looking at the camera. & a cat looking at the camera. & \bf 3.42 & 4.13 & 0.8 \\
2 &  he also said shockey is ``living the dream life of a new york athlete. & ``jeremy's a good guy,'' barber said, adding:``jeremy is living the dream life of the new york athlete. & \bf 3.55 &  4.22 &  2.75 \\
\hline
3 & bloomberg chips in a billion & bloomberg gives \$1.1 b to university & \bf 3.99 & 3.04 & 4.0 \\
4 & in other regions, the sharia is imposed. & in other areas, sharia law is being introduced by force. & \bf 4.44 & 3.72 & 4.75 \\
\hline
5 & three men in suits sitting at a table. & two women in the kitchen looking at a object. & 3.33 & \bf 2.79 & 0.0\\
6 & we never got out of it in the first place! & where does the money come from in the first place?	& 4.00	& \bf 3.33 & 0.8\\
7 & two birds interacting in the grass. & two dogs play with each other outdoors.	& 3.44	& \bf 2.81 & 0.2\\
\hline
\end{tabular}
\caption{
Illustrative sentence pairs from the STS datasets showing  errors made by \lstmavg and \wordavg. 
The last three columns show the gold similarity score, the similarity score of \lstmavg, and the similarity score of \wordavg. Boldface indicates smaller error compared to gold scores.
}
\label{table:errors}
\end{table*}

We also experiment with initializing the supervised models using our pretrained sentence model parameters, for the \wordavg model (no regularization), \lstmavg (dropout, scrambling), and \ganlong (dropout, scrambling) models from Table~\ref{table:transfer} and Table~\ref{table:gan}. We both initialize and then regularize back to these initial values, referring to this setting as ``universal''.\footnote{In these experiments, we tuned $\lambda_w$ over $\{10,1,10^{-1},10^{-2},10^{-3},10^{-4},10^{-5},10^{-6},10^{-7},10^{-8},0\}$ and $\lambda_c$ over $\{10,1,10^{-1},10^{-2},10^{-3},10^{-4},10^{-5},10^{-6},0\}$.} 

\begin{table}[t]
\setlength{\tabcolsep}{4pt}
\small
\centering
\begin{tabular} {| l | l | c | c |} \hline
Model & Regularization & STS & SICK \\
\hline
\multirow{2}{*}{\wordavg}
 & dropout & 80.7 & 84.5  \\
 & dropout, universal & \bf 82.9 & 85.6  \\
\hline
\multirow{2}{*}{\lstmavg}
 & dropout, scrambling & 76.5 & 84.0 \\
 & dropout, scrambling, universal & 81.3 & 85.2 \\
\hline
\multirow{2}{*}{\gan}
 & dropout, scrambling & 81.6 & 85.1 \\
 & dropout, scrambling, universal & 82.7 & \bf 86.0 \\
\hline
\end{tabular}
\caption{\label{table:universal} 
Impact of initializing and regularizing toward universal models (Pearson's $r \times 100$) in supervised training.
}
\end{table}

The results are shown in Table~\ref{table:universal}. 
Initializing and regularizing to the pretrained models significantly improves the performance for all three models, justifying our claim that these models serve a dual purpose: they can be used a black box semantic similarity function, and they possess rich knowledge that can be used to improve the performance of downstream tasks.

\section{Analysis} 
\subsection{Error Analysis}

We analyze the predictions of \wordavg and the recurrent networks, represented by \lstmavg, on the 20 STS datasets. We choose \lstmavg as it correlates slightly less strongly with \wordavg than the \gan on the results over all SemEval datasets used for evaluation. 
We scale the models' cosine similarities 
to lie within $[0,5]$, then 
compare the predicted similarities of \lstmavg and \wordavg to the gold similarities. We analyzed instances in which each model would tend to overestimate or underestimate the gold similarity relative to the other. These are illustrated in Table~\ref{table:errors}.

We find that \wordavg tends to overestimate the semantic similarity of a sentence pair, relative to \lstmavg, when the two sentences have a lot of word or synonym overlap, but have either important differences in key semantic roles or where one sentence has significantly more content than the other. These phenomena are shown in examples 1 and 2 in Table~\ref{table:errors}. Conversely, \wordavg tends to underestimate similarity when there are one-word-to-multiword paraphrases between the two sentences as shown in examples 3 and 4. 

\lstmavg tends to overestimate similarity when the two inputs have similar sequences of syntactic categories, 
but the meanings of the sentences are different (examples 5, 6, and 7). Instances of \lstmavg underestimating the similarity relative to \wordavg are relatively rare, and those that we found did not have any systematic patterns.

\subsection{\gan Gate Analysis}

We also investigate what is learned by the  gating function of the \ganlong. We are interested to see whether its estimates of importance correlate with those of traditional syntactic and (shallow) semantic analysis. 

We use the oracle trained \ganlong from Table~\ref{table:gan} and calculate the L$_1$ norm of the gate after embedding 
10,000 sentences from English Wikipedia.\footnote{We selected only sentences of less than or equal to 15 tokens to ensure more accurate parsing.}
We also automatically tag and parse these sentences using the Stanford dependency parser~\cite{manning-EtAl:2014:P14-5}. We then compute the average gate L$_1$ norms for particular part-of-speech tags, dependency arc labels, and their conjunction.

\begin{table}[t]
\setlength{\tabcolsep}{4pt}
\small
\centering
\begin{tabular} {|l|l|l|l|} \hline
\multicolumn{2}{|c|}{POS} & \multicolumn{2}{c|}{Dep. Label} \\
\hline
top 10 & bot. 10 & top 10 & bot. 10 \\
\hline
NNP & TO & number & possessive \\
NNPS & WDT & nn & cop \\
CD & POS & num & det \\
NNS & DT & acomp & auxpass \\
VBG & WP & appos & prep \\
NN & IN & pobj & cc \\
JJ & CC & vmod & mark \\
UH & PRP & dobj & aux \\
VBN & EX & amod & expl \\
JJS & WRB & conj & neg \\
\hline
\end{tabular}
\caption{\label{table:qa} 
POS tags and dependency labels with highest and lowest average \ganlong gate L$_1$ norms. The lists are ordered from highest norm to lowest in the top 10 columns, and lowest to highest in the bottom 10 columns.
}
\end{table}

Table~\ref{table:qa} shows the highest/lowest average norm tags and dependency labels. The network prefers nouns, especially proper nouns, as well as cardinal numbers, which is sensible as these are among the most discriminative features of a sentence.

Analyzing the dependency relations, we find that nouns in the object position tend to have higher weight than nouns in the subject position. 
This may relate to topic and focus; the object may be more likely to be the ``new'' information related by the sentence, which would then make it more likely to be matched by the other sentence in the paraphrase pair. 

\begin{table}
\setlength{\tabcolsep}{4pt}
\centering
\begin{tabular} {|l|l|} 
\hline
Dep. Label & Weight \\
\hline
xcomp & 170.6 \\
acomp & 167.1 \\
root & 157.4 \\
\hline
amod & 143.1 \\
\hline
advmod & 121.6 \\
\hline
\end{tabular}
\caption{\label{table:jj} 
Average L$_1$ norms for adjectives (JJ) with selected  dependency labels.
}
\end{table}

\begin{table}
\setlength{\tabcolsep}{4pt}
\centering
\begin{tabular} {|l|l|} 
\hline
Dep. Label & Weight \\
\hline
pcomp & 190.0 \\
amod & 178.3 \\
xcomp & 176.8 \\
\hline
vmod & 170.6 \\
root & 161.8 \\
\hline
auxpass & 125.4 \\
prep & 121.2 \\
\hline
\end{tabular}
\caption{\label{table:vbg} 
Average L$_1$ norms for words with the tag VBG with selected dependency labels. 
}
\end{table}

We find that the weights of adjectives depend on their position in the sentence, as shown in Table~\ref{table:jj}. The highest norms appear when an adjective is an xcomp, acomp, or root; this typically means it is residing in an object-like position in its clause. Adjectives that modify a noun (amod) have medium weight, and those that modify another adjective or verb (advmod) have low weight.

Lastly, we analyze words tagged as VBG, a highly ambiguous tag that can serve many syntactic roles in a sentence. As shown in Table~\ref{table:vbg}, we find that when they are used to modify a noun (amod) or in the object position of a clause (xcomp, pcomp) they have high weight. Medium weight appears when used in verb phrases (root, vmod) and low weight when used as prepositions or auxiliary verbs (prep, auxpass).

\section{Conclusion}
We showed how to modify and regularize LSTMs to improve their performance for learning paraphrastic sentence embeddings in both transfer and supervised settings. 
We also introduced a new recurrent network, the \ganlong, that improves upon both \wordavg and LSTMs for these tasks, and we release our code and trained models. 

Furthermore, we analyzed the different errors produced by \wordavg and the recurrent methods and found that the recurrent methods were learning composition that wasn't being captured by \wordavg. We also investigated the \gan in order to better understand the compositional phenomena it was learning by analyzing the L$_1$ norm of its gate over various inputs.

Future work will explore additional data sources, including  
from aligning different translations of novels~\cite{barzilay2001extracting}, aligning new articles of the same topic~\cite{dolan2004unsupervised}, or even possibly using machine translation systems to translate bilingual text into paraphrastic sentence pairs. Our new techniques, combined with the promise of new data sources, 
offer a great deal of potential for improved universal paraphrastic sentence embeddings.

\section*{Acknowledgments}

We thank the anonymous reviewers for their valuable comments. This research used resources of the Argonne Leadership Computing Facility, which is a DOE Office of Science User Facility supported under Contract DE-AC02-06CH11357. We thank the developers of Theano~\cite{2016arXiv160502688short} and NVIDIA Corporation for donating GPUs used in this research.

\bibliography{acl2017}
\bibliographystyle{acl_natbib}

\end{document}